\documentclass[11pt,a4paper]{article}

\usepackage[T1]{fontenc}
\usepackage[utf8]{inputenc}
\usepackage[english]{babel}

\usepackage{mathpazo} %
\usepackage[scaled=0.92]{helvet} %
\usepackage{courier} %
\usepackage{amsmath}
\usepackage{microtype} %

\usepackage{geometry}
\usepackage{setspace}
\usepackage{graphicx}
\usepackage{xcolor}
\definecolor{arxivblue}{rgb}{0.0, 0.22, 0.66}
\definecolor{orcidlogocol}{HTML}{A6CE39}

\usepackage{booktabs}
\usepackage{float}
\usepackage[section]{placeins}
\usepackage{caption}
\usepackage{subcaption}
\usepackage{titlesec}
\titleformat{\section}{\Large\sffamily\bfseries}{\thesection}{1em}{}
\titleformat{\subsection}{\large\sffamily\bfseries}{\thesubsection}{1em}{}

\usepackage{csquotes}
\usepackage[backend=biber,style=numeric,sorting=none]{biblatex}
\usepackage{hyperref}
\usepackage{fontawesome5}
\hypersetup{
    colorlinks=true,
    linkcolor=arxivblue,
    citecolor=arxivblue,
    urlcolor=arxivblue,
    pdftitle={Text-as-Signal},
    pdfauthor={Hugo Moreira}
}

\makeatletter
\def\@maketitle{%
  \newpage
  \null
  \vskip 1.5em%
  \begin{center}%
  \let \footnote \thanks
    \hrule height 1.5pt %
    \vskip 0.8em
    {\LARGE \rmfamily \bfseries \@title \par}%
    \vskip 0.8em
    \hrule height 1.5pt %
    \vskip 1.5em%
    {\large
      \lineskip .5em%
      \begin{tabular}[t]{c}%
        \@author
      \end{tabular}\par}%
  \end{center}%
  \par
  \vskip 1.5em}
\makeatother

\title{Text-as-Signal:\\Quantitative Semantic Scoring with Embeddings, Logprobs, and Noise Reduction}
\author{%
  \href{https://orcid.org/0000-0002-4199-6006}{\textcolor{orcidlogocol}{\faOrcid}\hspace{1mm}\textcolor{black}{\sffamily\textbf{Hugo Moreira}}} \\[0.3em]
  \normalsize \sffamily ISCTE-IUL, Portugal\\[0.3em]
  \small \texttt{hugo\_filipe\_moreira@iscte-iul.pt}
}
\date{}

\begin{document}

\maketitle

\begin{abstract}
This paper presents a practical pipeline for turning text corpora into quantitative semantic signals. Each news item is represented as a full-document embedding, scored through logprob-based evaluation over a configurable positional dictionary, and projected onto a noise-reduced low-dimensional manifold for structural interpretation. In the present case study, the dictionary is instantiated as six semantic dimensions and applied to a corpus of 11,922 Portuguese news articles about Artificial Intelligence. The resulting identity space supports both document-level semantic positioning and corpus-level characterization through aggregated profiles. We show how Qwen embeddings, UMAP, semantic indicators derived directly from the model output space, and a three-stage anomaly-detection procedure combine into an operational text-as-signal workflow for AI engineering tasks such as corpus inspection, monitoring, and downstream analytical support. Because the identity layer is configurable, the same framework can be adapted to the requirements of different analytical streams rather than fixed to a universal schema.
\end{abstract}

\section{Introduction}

Dense embeddings are effective for representing documents, but raw vector spaces are difficult to use on their own in operational settings. To treat text as an operational signal, the semantic coordinates within the corpus must be extracted, structured, and expressed as continuous variables on bounded scales. Once established, these quantitative scores become directly usable for downstream AI engineering tasks such as aggregation, monitoring, regression, and threshold-based routing, without requiring a human analyst to interpret the latent space indirectly.

An underlying premise of our approach is that an interaction with a Large Language Model (LLM) does not have to mimic explicit human communication. Instead, a model's weights can, and should, be treated as a highly compressed topology of human language. This work takes that idea to its limit by intentionally bypassing traditional generative content analysis. Rather than prompting the model to generate explicit, unstructured text labels, we treat its output space strictly as an evaluator of latent linguistic signals.

This paper describes a pipeline for extracting these semantic signals from a text corpus. The workflow starts from full-document embeddings, computes logprob-based semantic indicators directly from each news item, and stabilizes the data structure by projecting it into a lower-dimensional manifold and applying a three-step anomaly-detection process for noise reduction. The result is a fully operationalized semantic space that supports pattern recognition, corpus inspection, and continuous semantic scoring of documents.

In this study, the unit of analysis is the news item itself rather than a retrieval-oriented chunk or passage. Each article receives a continuous semantic identity across six dimensions: opportunity vs. risk, regulatory pressure, economic momentum, ethics vs. utility, geopolitical scope, and urgency. Once assigned at the article level, that same identity dictionary can be aggregated to characterize the corpus as a whole, individual clusters, or temporal windows when longitudinal analysis is needed, producing positional profiles rather than only a collection of isolated document scores. Because the identity is continuous, it can also support downstream regression and learning tasks, making it possible to test whether particular semantic regions are associated with stronger outcomes.

We therefore frame text-as-signal processing not merely as a visual exercise in latent space mapping, but as a concrete capability in a modern AI data pipeline. The contribution of this paper is to document that capability in a focused AI engineering workflow and to show, on a corpus of Portuguese AI news, how quantitative document-level semantic identity and corpus-level characterization can be successfully extracted inside the same pipeline. %
\section{Related Work}

The task of semantic map generation connects text representation learning with topological data analysis. Early approaches to corpus mapping relied heavily on term-frequency matrices and probabilistic modeling, such as Latent Dirichlet Allocation (LDA), to extract thematic structures. The transition towards dense embeddings has largely superseded those methods. Frameworks such as BERTopic~\parencite{grootendorst2022bertopic} have demonstrated that clustering sentence representations (typically via UMAP~\parencite{mcinnes2018umap} and HDBSCAN) followed by term-weighting modifications yields highly coherent semantic regions. 

Simultaneously, Large Language Models (LLMs) have emerged as powerful zero-shot text annotators. While many pipelines rely on generative extraction (prompting for free-text labels), recent work demonstrates that directly reading the model's output distribution---specifically through zero-shot logprob expectations---provides a more stable, calibrated, and continuous signal for classification tasks~\parencite{wallace2024meandifference}. 

Our pipeline bridges these two parallel tracks. It adopts the structural clustering paradigm of dense topological methods to build the core map and conduct noise reduction, while replacing discrete term-based descriptors with continuous, LLM-scored logprob coordinates to form an interpretable positional dictionary.
\section{Method}

The pipeline has four stages.

First, each news article is treated as a single semantic unit and embedded with the Qwen2.5 8B Instruct model, producing 4096-dimensional vectors. No chunking is applied, since the analytical unit is the news item itself rather than a retrieval passage. The model was selected because it offered strong open-weight performance on text-embedding benchmarks and competitive clustering-oriented results on the public MTEB evaluation ecosystem~\parencite{muennighoff2023mteb,mtebLeaderboard}, making it a pragmatic choice for structure-preserving document representation. Embedding generation is executed through \texttt{vLLM} and the resulting vectors are stored in PostgreSQL with \texttt{pgvector} support.

Second, the embedding space is reduced with UMAP~\parencite{mcinnes2018umap} into a 5D latent representation for structural analysis and a 2D projection for visualization. The 5D choice is aligned with the estimated intrinsic dimensionality of the corpus, measured at approximately $d \approx 4.11$ using TwoNN~\parencite{facco2017estimating}. K-Means with $K=15$ is then applied to the 5D manifold as the initial structural partition. The value of $K$ was chosen as a practical resolution parameter: lower values merged semantically distinct regions, whereas higher values produced partitions that were too fragmented for stable interpretation. In this corpus, $K=15$ yielded readable regions with roughly 800 articles per cluster on average. After anomaly detection and noise reduction, the retained core map corresponds to a trimmed 13-region structural solution. HDBSCAN is retained only as an exploratory density diagnostic.

\begin{figure}[htbp]
\centering
\includegraphics[width=0.60\linewidth]{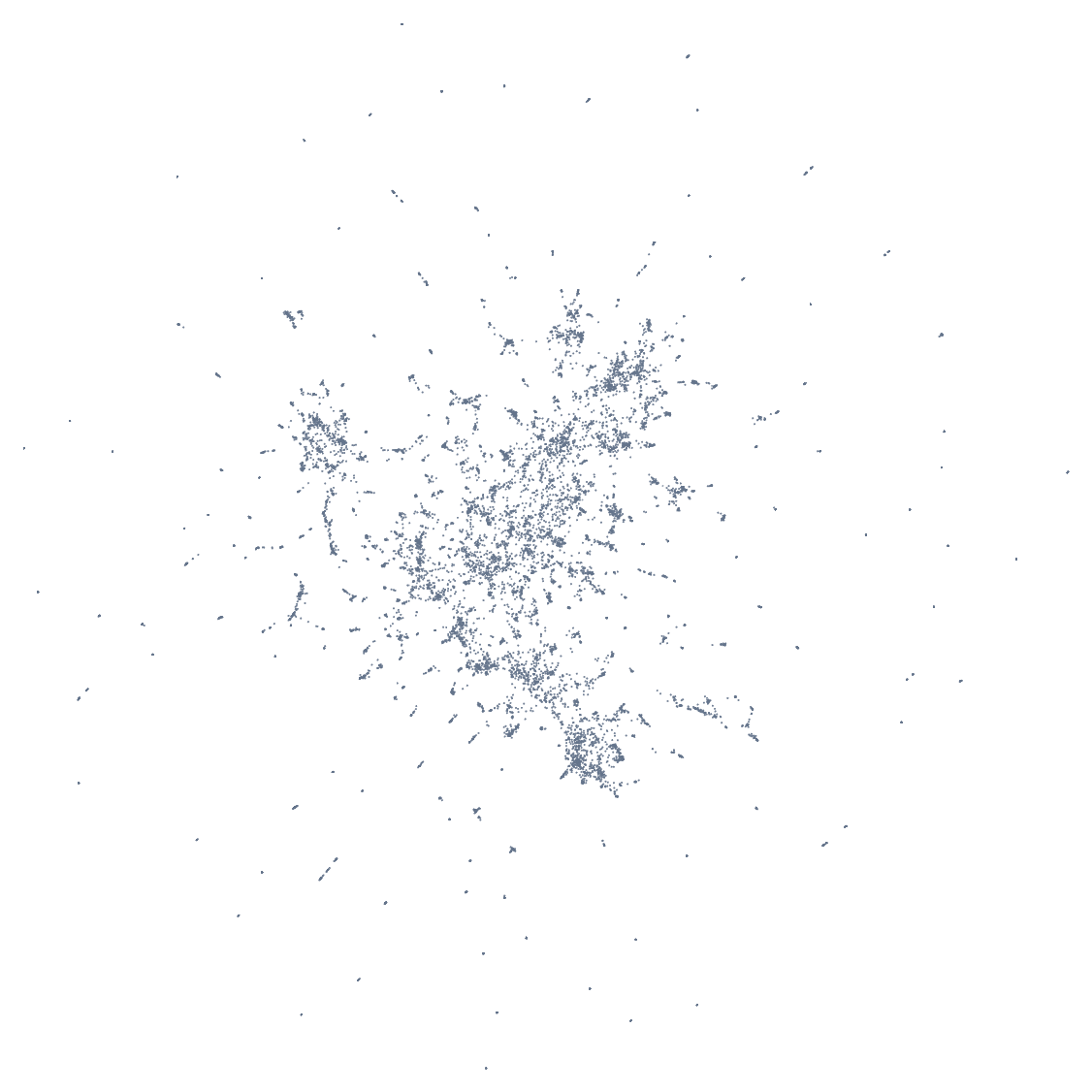}
\caption{Base 2D semantic topography obtained from the UMAP projection of the full-document embedding space.}
\label{fig:short_umap_base}
\end{figure}

Third, each news article is evaluated against a fixed set of semantic targets through logprob-based zero-shot scoring, producing continuous semantic indicators in the interval $[0,1]$. Instead of generating free text labels, the model output space is queried directly from the article input: the final hidden state is compared against target label vectors, transformed through a softmax layer, and converted into a continuous expectation score~\parencite{wallace2024meandifference}. This produces a fixed positional dictionary with six dimensions: \textit{Opportunity vs. Risk}, \textit{Regulatory Pressure}, \textit{Economic Momentum}, \textit{Ethics vs. Utility}, \textit{Geopolitical Scope}, and \textit{Urgency}. These article-level indicators provide semantic coordinates that complement the geometric structure of the embedding manifold. Once computed, they are painted onto the 2D projection so that the same semantic identity can be read spatially across the map.

To make the pipeline explicit, let the corpus be $\mathcal{D} = \{d_1, \ldots, d_n\}$, where each $d_i$ is a full news article. The embedding model defines a document encoder
\[
x_i = f_{\mathrm{enc}}(d_i), \qquad x_i \in \mathbb{R}^{4096}.
\]
The geometric structure is then derived through two UMAP projections,
\[
z_i = g_5(x_i), \qquad y_i = g_2(x_i),
\]
where $z_i \in \mathbb{R}^{5}$ is used for structural partitioning and $y_i \in \mathbb{R}^{2}$ for visualization. The initial region assignment is obtained by
\[
c_i = \operatorname{KMeans}_{K=15}(z_i).
\]

For each semantic dimension $m$, defined by an ordered pole pair $(\ell_m^{-}, \ell_m^{+})$, the model returns article-conditioned log-scores $\lambda_{i,m}^{-}$ and $\lambda_{i,m}^{+}$. These are converted into a normalized article-level indicator
\[
s_{i,m} = \frac{\exp(\lambda_{i,m}^{+})}{\exp(\lambda_{i,m}^{-}) + \exp(\lambda_{i,m}^{+})}, \qquad s_{i,m} \in [0,1],
\]
so that values near $0$ indicate affinity with the lower pole and values near $1$ indicate affinity with the upper pole. The same mechanism can be used for a single semantic target $\tau$ as a centrality probe, yielding a target-specific score $s_{i,\tau}$ for each article.

Noise reduction is applied as a sequence of explicit filters over the 2D topography. Let $P \subseteq \mathcal{D}$ denote the subset of articles that belong to at least one HDBSCAN cluster, excluding points labeled as density noise. The global centroid of this structural core is
\[
C_{\mathrm{global}} = \left( \frac{1}{|P|}\sum_{d_i \in P} y_{i1},\; \frac{1}{|P|}\sum_{d_i \in P} y_{i2} \right).
\]
For any article $d_i$, define its topographic distance to the continental center by
\[
d_i^{\mathrm{global}} = \lVert y_i - C_{\mathrm{global}} \rVert_2.
\]
Let $\mu_{\mathrm{global}}$ and $\sigma_{\mathrm{global}}$ be the mean and standard deviation of these distances computed over the full projected dataset. The global outlier filter is then
\[
G(i) = \mathbf{1}\!\left[d_i^{\mathrm{global}} \le \mu_{\mathrm{global}} + 1.2\,\sigma_{\mathrm{global}}\right].
\]
This separates articles that lie outside the semantic continent from points that HDBSCAN marks as density noise but which still remain inside the broad continental radius. In implementation terms, the centroid is anchored on the HDBSCAN core, while the radial threshold itself is calibrated against the full point cloud.

Within each K-Means region $K$, let
\[
C_K = \frac{1}{|P_K|}\sum_{d_i \in P_K} y_i
\]
be the regional centroid in the 2D topography, where $P_K$ is the set of articles assigned to that region. The local distance of article $d_i \in P_K$ is
\[
d_i^{\mathrm{local}} = \lVert y_i - C_K \rVert_2.
\]
If $\mu_K$ and $\sigma_K$ are the mean and standard deviation of these within-region distances, the local-maverick filter is
\[
L(i) = \mathbf{1}\!\left[d_i^{\mathrm{local}} \le \mu_K + 1.8\,\sigma_K\right].
\]
This second pass removes points that are not globally isolated but remain anomalous relative to their own semantic region.

The final pass removes structurally disconnected semantic islands. For a chosen topological grouping, define an undirected graph $G_\varepsilon = (V,E)$ with one vertex per article and edges given by
\[
(u,v) \in E \iff \lVert u - v \rVert_2 < \varepsilon.
\]
We then extract the connected components $C_1, C_2, \ldots, C_r$ ordered by size, so that $|C_1| \ge |C_2| \ge \cdots \ge |C_r|$. Articles are retained only if they belong to the largest connected component. In practice, $\varepsilon = 1.2$ for the K-Means graph and $\varepsilon = 1.0$ for the HDBSCAN diagnostic graph. Denoting this reachability test by $R(i)$, an article is retained in the final semantic map only if
\[
\chi_i = G(i)\,L(i)\,R(i) = 1.
\]
The final map can therefore be written as
\[
\mathcal{M} = \{(y_i, c_i, s_i) \mid d_i \in \mathcal{D},\; \chi_i = 1\},
\]
where $s_i$ denotes the vector of semantic scores attached to article $d_i$. This makes the pipeline analytically precise: embeddings provide document geometry, logprob scoring provides semantic coordinates, and noise reduction defines the stable subset on which the final interpretation is based.

\begin{table}[H]
\centering
\small
\caption{Semantic identity dimensions used in the positional dictionary.}
\label{tab:short_dimensions}
\begin{tabular}{p{0.38\linewidth}p{0.25\linewidth}p{0.25\linewidth}}
\toprule
\textbf{Dimension} & \textbf{Low Pole} & \textbf{High Pole} \\
\midrule
Opportunity vs. Risk & Opportunity & Danger \\
Regulatory Pressure & Deregulation & Compliance \\
Economic Momentum & Niche & Commercial \\
Ethics vs. Utility & Human-centric & Efficiency \\
Geopolitical Scope & Local / EU & Global \\
Urgency & Analytical & Breaking / Alarmist \\
\bottomrule
\end{tabular}
\end{table}

Fourth, the projected map is refined through three anomaly-detection steps. Global outliers are detected through distance to the HDBSCAN-defined continental centroid in the 2D topography using a 1.2$\sigma$ threshold\parencite{ruff2018deep}; local mavericks are detected inside each K-Means region using a 1.8$\sigma$ threshold; and a graph-based reachability step, inspired by SCAN~\parencite{xu2007scan}, removes structurally disconnected islands through connected-component pruning. The resulting filtered topology defines the final semantic map.

\begin{figure}[htbp]
\centering
\begin{subfigure}[t]{0.48\linewidth}
\centering
\includegraphics[width=\linewidth]{images/theseus-topography-umap-base.png}
\caption{Baseline topography.}
\end{subfigure}
\hfill
\begin{subfigure}[t]{0.48\linewidth}
\centering
\includegraphics[width=\linewidth]{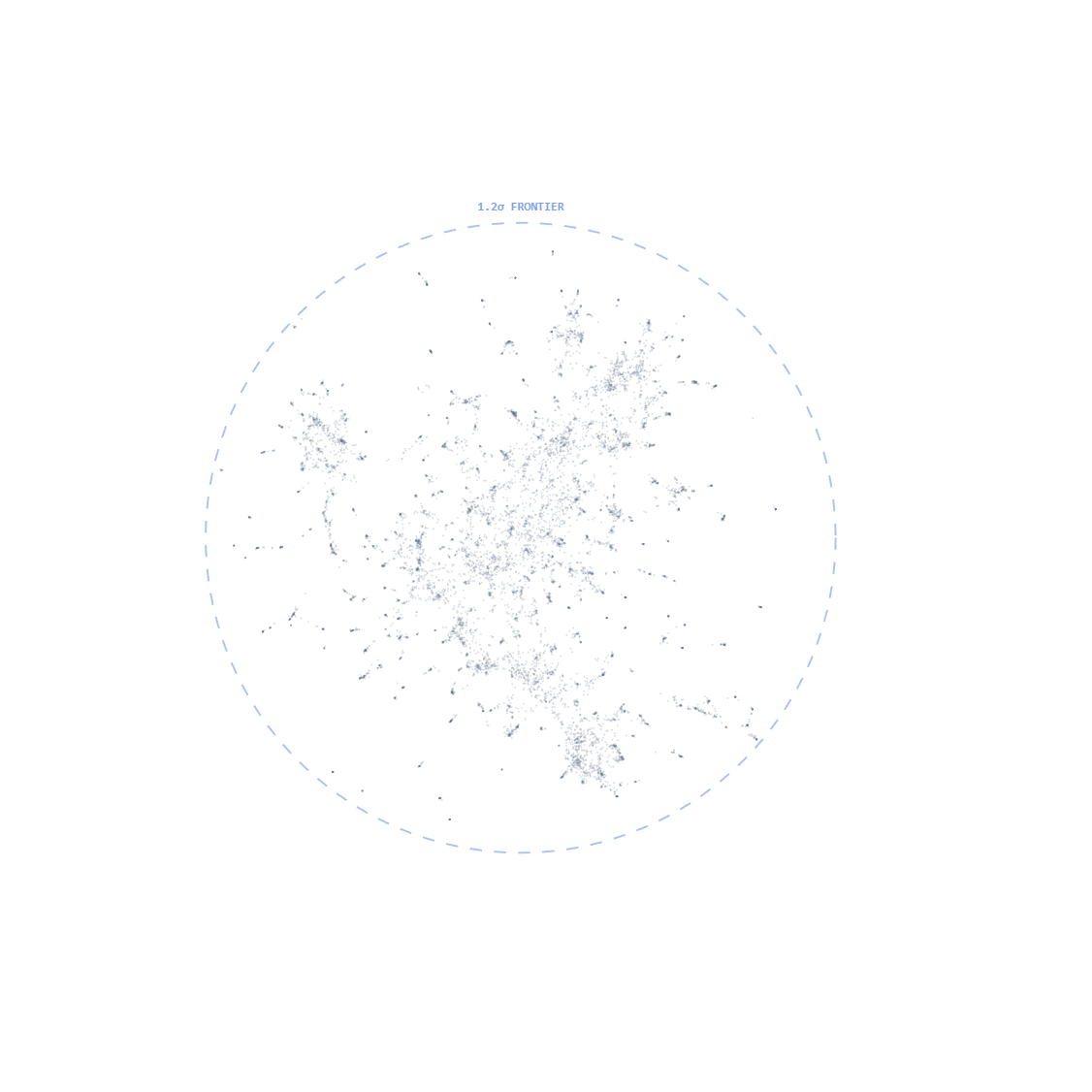}
\caption{Global distance pass.}
\end{subfigure}
\\[1em]
\begin{subfigure}[t]{0.48\linewidth}
\centering
\includegraphics[width=\linewidth]{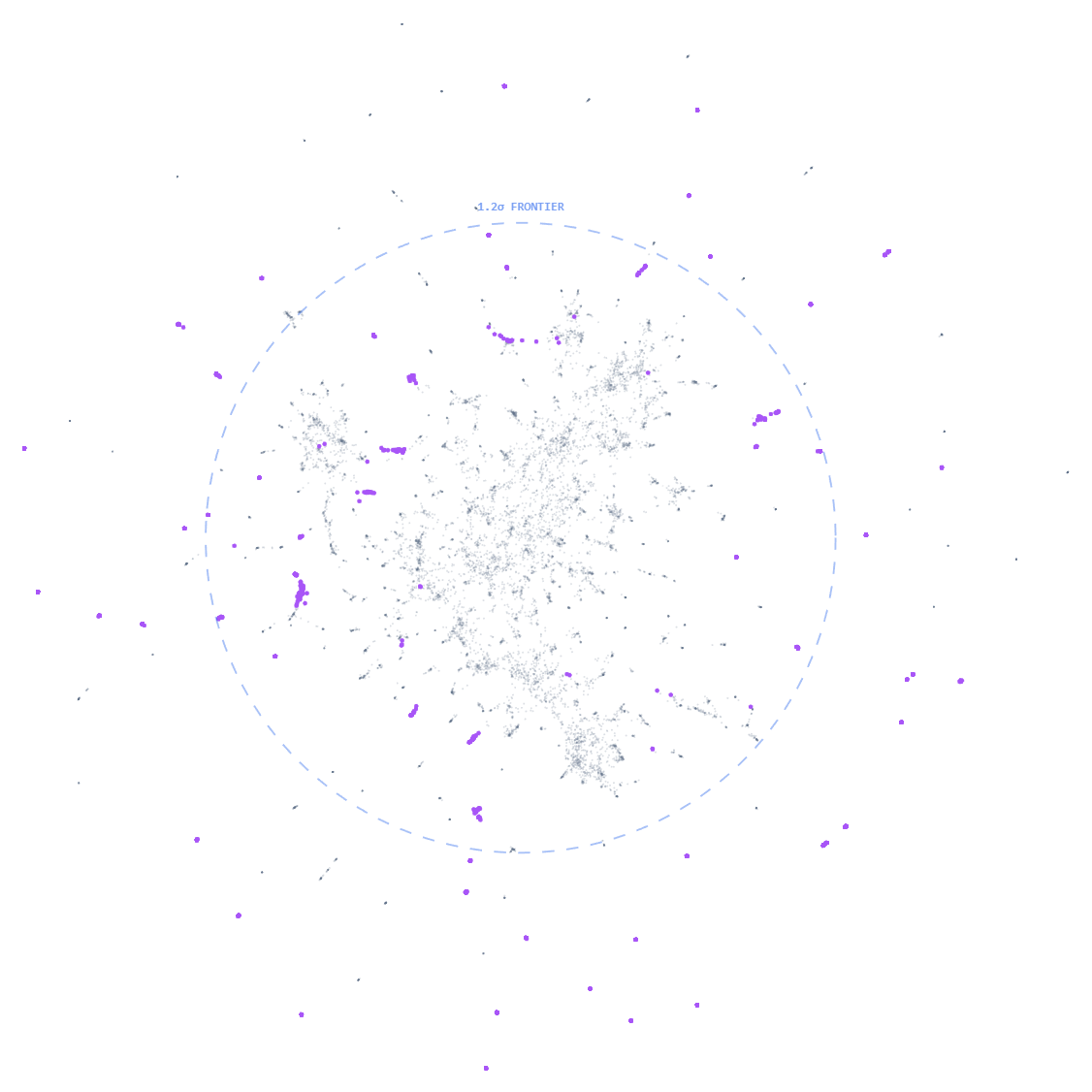}
\caption{Local outlier pass.}
\end{subfigure}
\hfill
\begin{subfigure}[t]{0.48\linewidth}
\centering
\includegraphics[width=\linewidth]{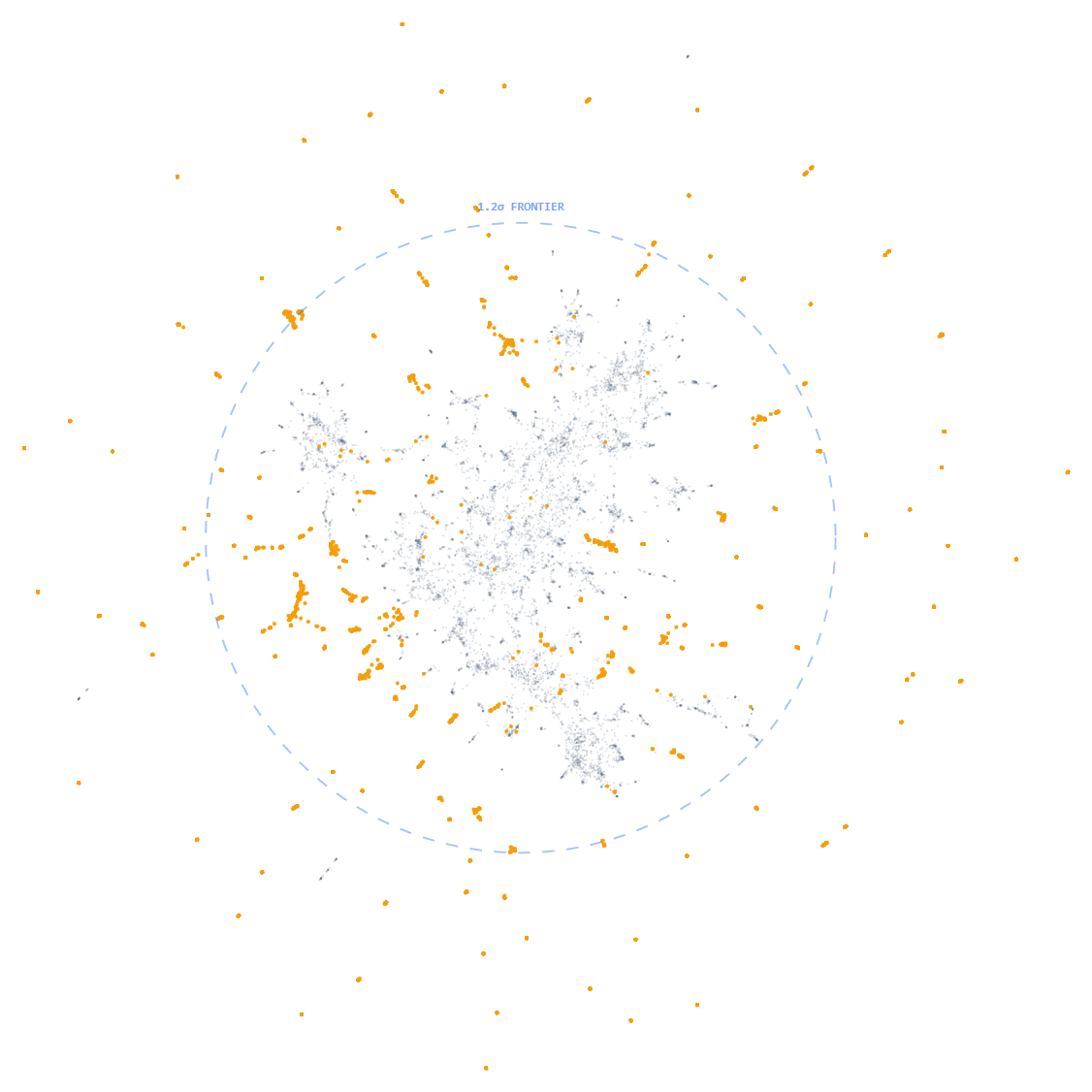}
\caption{Structural outlier pass.}
\end{subfigure}
\caption{Anomaly-detection sequence used for noise reduction. The first panel repeats the baseline semantic topography, followed by the global distance pass, the local outlier pass, and the structural outlier pass.}
\label{fig:short_anomaly_pipeline}
\end{figure}

The same positional dictionary is also used in aggregate. After scoring the articles individually, we summarize the distribution of the corpus across fixed quartile bands $[0,0.25]$, $(0.25,0.50]$, $(0.50,0.75]$, and $(0.75,1]$, yielding a corpus-level semantic profile that can be read alongside the topographic map and the integrated semantic identity view shown later in Table~\ref{tab:short_semantic_identity_overview}. For centrality and related continuous indicators, the backend also computes descriptive statistics such as mean, continuous median, sample standard deviation, and outlier proportions derived from absolute $Z$-scores, which are used for profiling rather than for the topographic pruning step. %
\section{Case Study}

We apply the pipeline to a corpus of 11,922 Portuguese news articles related to Artificial Intelligence. The dataset spans 2022 to 2024, contains no null values, and includes only records that mention \textit{Intelig\^encia artificial} and/or \textit{AIAct} in either the title or the description. The working textual fields are the title and description, with the latter providing a substantial article-level summary rather than a short retrieval snippet.

The corpus is large enough to expose both stable thematic regions and a substantial semantic periphery. Descriptive profiling shows a mean character count of 6518.1 characters (median 3937.5; standard deviation 9389.0) and a mean of 2.05 AI-related mentions per item (median 1.0; standard deviation 2.83). This is useful in practice because the pipeline has to handle both concise headlines and long-form articles without changing the analytical unit.

The case study therefore provides a realistic setting in which semantic regions, semantic poles, anomalous documents, and corpus-level semantic tendencies can be observed simultaneously. Because the scoring is attached to each article, the pipeline does not only cluster the corpus: it gives each news item a readable semantic position and then uses those same positions to summarize the character of the full collection. In that sense, the corpus also acts as a practical satellite for tasks that are often underemphasized in data engineering and data science workflows, where text is treated not only as narrative content but as an operational signal for monitoring, segmentation, anomaly detection, and downstream decision support. %
\section{Findings}

The resulting map shows that geometric structure alone is not sufficient for interpretation. The pipeline first has to resolve a readable structural partition and remove topological instability; only then do the logprob-based semantic signals become fully legible across the corpus.

\subsection{Structural Resolution and Noise Reduction}

At the structural level, the initial K-Means segmentation yields a 15-region map of the Portuguese AI news landscape. This partition reveals a fragmented semantic space in which stable narrative regions coexist with a broad periphery of transitional or niche reports. After anomaly detection is applied as part of the noise-reduction process, two of those regions are no longer retained as stable parts of the core map, leaving a trimmed 13-region solution for the final structural reading. HDBSCAN provides a useful density diagnostic, but it classified 5450 articles as noise, excluding roughly 45\% of the corpus and making it too restrictive for the final structural representation.

The anomaly-detection stage drives the noise-reduction process by identifying peripheral and structurally unstable observations. The three-step pipeline isolates 1282 global outliers, 944 local mavericks, and a final set of graph-level structural outliers, for a total of 2565 unique articles removed from the final core map. This corresponds to about 21.5\% of the corpus, leaving a substantially cleaner topology while preserving broad coverage of the discourse.

\begin{figure}[htbp]
\centering
\begin{subfigure}[t]{0.48\linewidth}
\centering
\includegraphics[width=\linewidth]{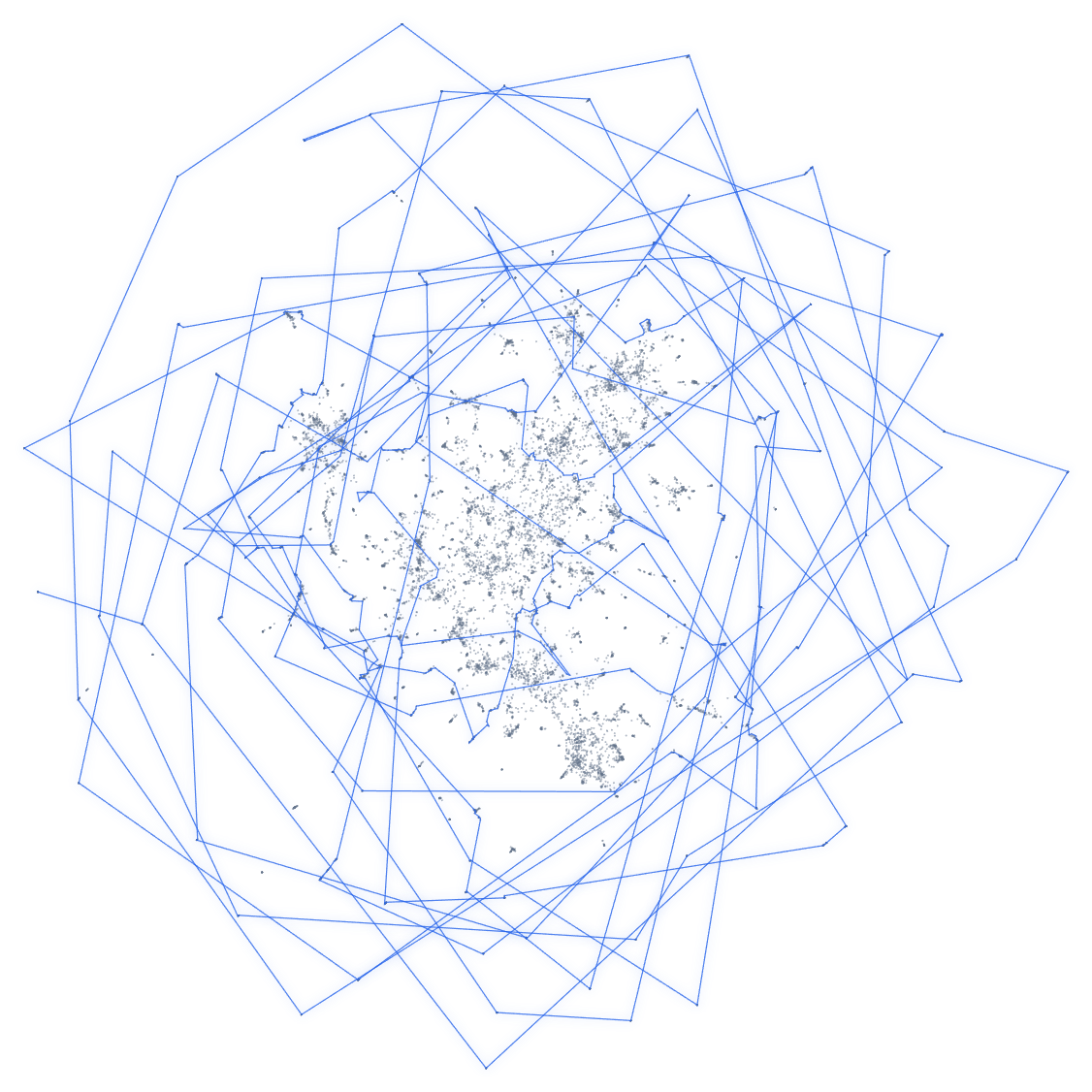}
\caption{K-Means with structural noise.}
\end{subfigure}
\hfill
\begin{subfigure}[t]{0.48\linewidth}
\centering
\includegraphics[width=\linewidth]{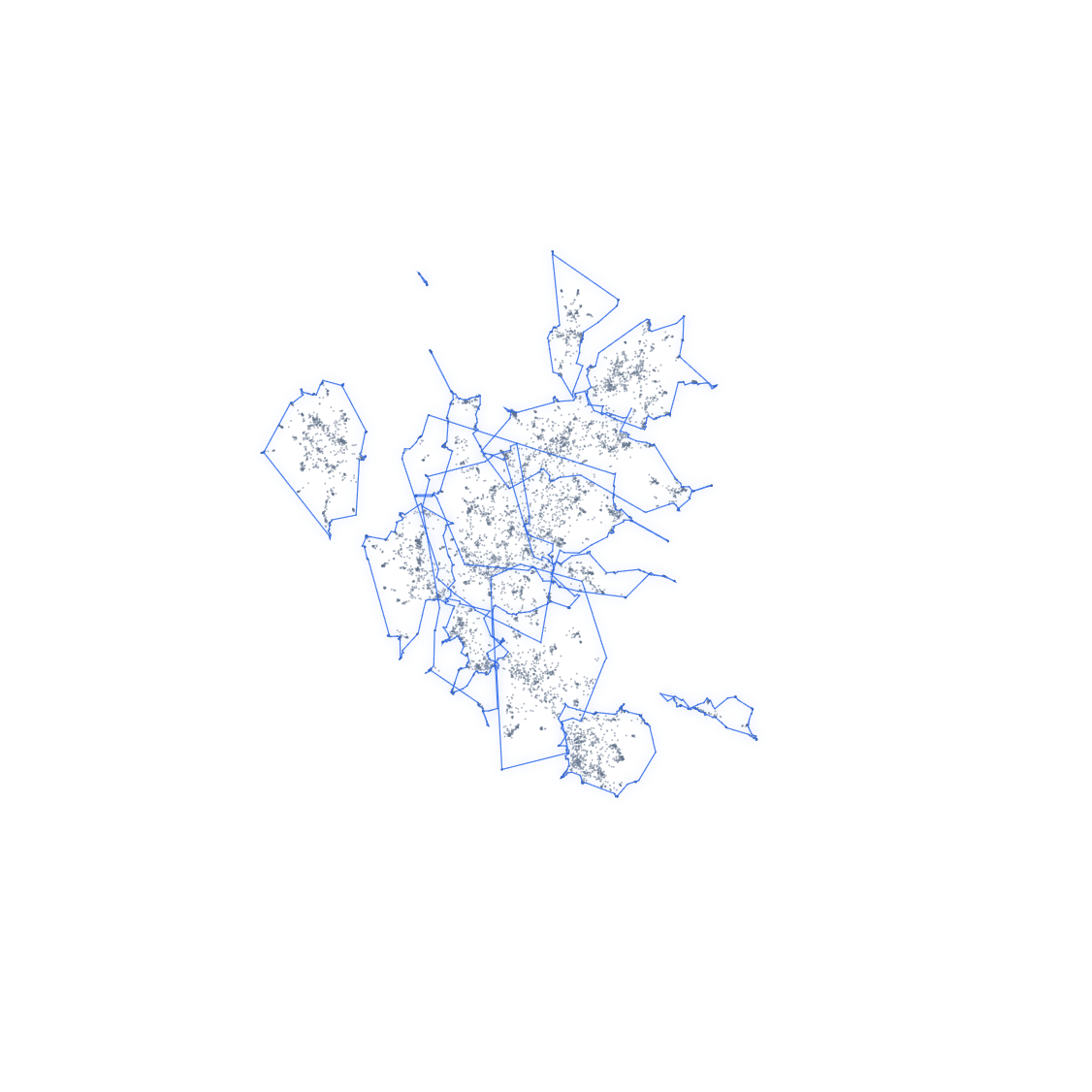}
\caption{K-Means after noise reduction.}
\end{subfigure}
\caption{K-Means regions before and after anomaly detection and noise reduction. The comparison shows why K-Means is central to the anomaly-detection analysis rather than only a downstream visualization step.}
\label{fig:short_kmeans_comparison}
\end{figure}

The trimmed map that remains after anomaly detection and noise reduction is not only easier to inspect visually. It also provides a more stable substrate for downstream tasks such as region-level characterization, semantic monitoring, anomaly review, and other forms of document-level or cluster-level analysis.

\subsection{Logprobs as Text Signal}

Once the structural map is stabilized, the logprob-based semantic signals provide the second reading of the corpus. Rather than functioning as decorative overlays, they turn the text itself into a measurable signal that can be used to probe topical alignment, describe semantic identity, and characterize both articles and aggregates.

The first identity layer is semantic centrality. For each article, a logprob-based score measures how strongly the text is centred on the AI topic. In practice, this acts as a semantic probe: it tests how the corpus aligns with the target object of interest before any richer identity dimensions are aggregated. In this case, the target object is Artificial Intelligence, but the same procedure could be applied to a specific company, political actor, institution, or event. It also functions as a quality-control instrument for corpus construction, because divergence between the observed centrality distribution and the expected target object becomes an early warning of mismatch between the collected corpus and the semantic object the analysis is supposed to represent.

\begin{figure}[htbp]
\centering
\includegraphics[width=1.0\linewidth]{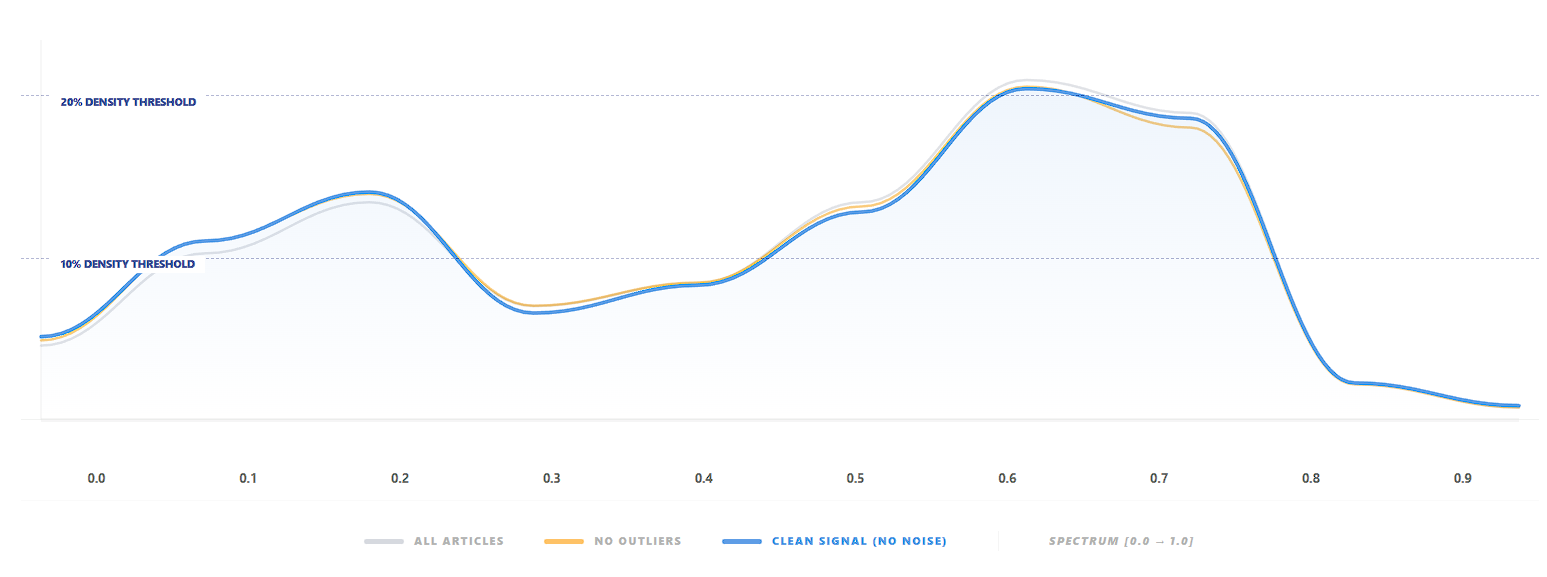}
\caption{Relative frequency distribution of semantic centrality scores. The plot compares all articles (grey) with the dataset after removing outliers (orange) and structural noise (blue), showing a prominent peak in the [0.6, 0.75] range.}
\label{fig:short_semantic_centrality}
\end{figure}

\noindent As illustrated in Fig.~\ref{fig:short_semantic_centrality}, the dataset exhibits a strong semantic centrality towards the theme "Artificial Intelligence". The frequency curve forms a distinct peak in the higher-score region, concentrated between 0.6 and 0.75, validating the effectiveness of the initial keyword-based filtering while preserving a measurable tail of lower-centrality items. Furthermore, the curves for the clean signal closely track the distribution of the full dataset. This confirms that the structural outliers removed during anomaly detection and noise reduction are not merely low-centrality items, but rather topologically unstable documents distributed across the entire semantic spectrum.

\begin{table}[htbp]
\centering
\caption{Aggregate positional dictionary view of the corpus. The table shows the quantitative distribution of articles across the quartiles of the six semantic identity dimensions, highlighting the dominant narrative poles.}
\label{tab:short_semantic_identity_overview}
\resizebox{\linewidth}{!}{%
\begin{tabular}{@{}lcccc@{}}
\toprule
\textbf{Dimension} & \textbf{0.00 -- 0.25 (Q1)} & \textbf{0.26 -- 0.50 (Q2)} & \textbf{0.51 -- 0.75 (Q3)} & \textbf{0.76 -- 1.00 (Q4)} \\
\midrule
\textbf{Opportunity vs. Risk} & Pure Opportunity & Growth Oriented & Risk Aware & Critical Danger \\
 & 3\% & \textbf{89\%} & 8\% & 0\% \\
\addlinespace
\textbf{Regulatory Pressure} & Deregulation & Low Supervision & Moderate Oversight & High Compliance \\
 & 25\% & \textbf{49\%} & 22\% & 3\% \\
\addlinespace
\textbf{Economic Momentum} & Academic/Niche & Emerging Market & Commercial Growth & Economic Engine \\
 & 3\% & 21\% & \textbf{67\%} & 10\% \\
\addlinespace
\textbf{Ethics vs. Utility} & Human-Centric & Balanced Ethics & Utility Focused & Max Efficiency \\
 & 4\% & \textbf{94\%} & 2\% & 0\% \\
\addlinespace
\textbf{Geopolitical Scope} & Local/Regional & National Scope & Continental/EU & Global/Interstate \\
 & \textbf{43\%} & 36\% & 14\% & 7\% \\
\addlinespace
\textbf{Urgency / Sentiment} & Educational & Analytical & Active News & Crisis/Breaking \\
 & 29\% & \textbf{63\%} & 7\% & 2\% \\
\bottomrule
\end{tabular}%
}
\end{table}

Beyond centrality, the positional dictionary extends semantic identity to six continuous dimensions. As illustrated in Table~\ref{tab:short_semantic_identity_overview}, at the aggregate level this system produces a readable profile of the corpus itself. In this case study, the collection heavily concentrates around opportunity-oriented, lower-supervision, economically expansionary, balanced-ethics, and analytically framed positions, while showing decidedly weaker occupancy in extreme risk, crisis, and hard-compliance bands. This matters because the entire corpus can be characterized through the same semantic language used to describe individual documents.

\begin{figure}[htbp]
\centering
\begin{subfigure}[t]{0.32\linewidth}
\centering
\includegraphics[width=\linewidth]{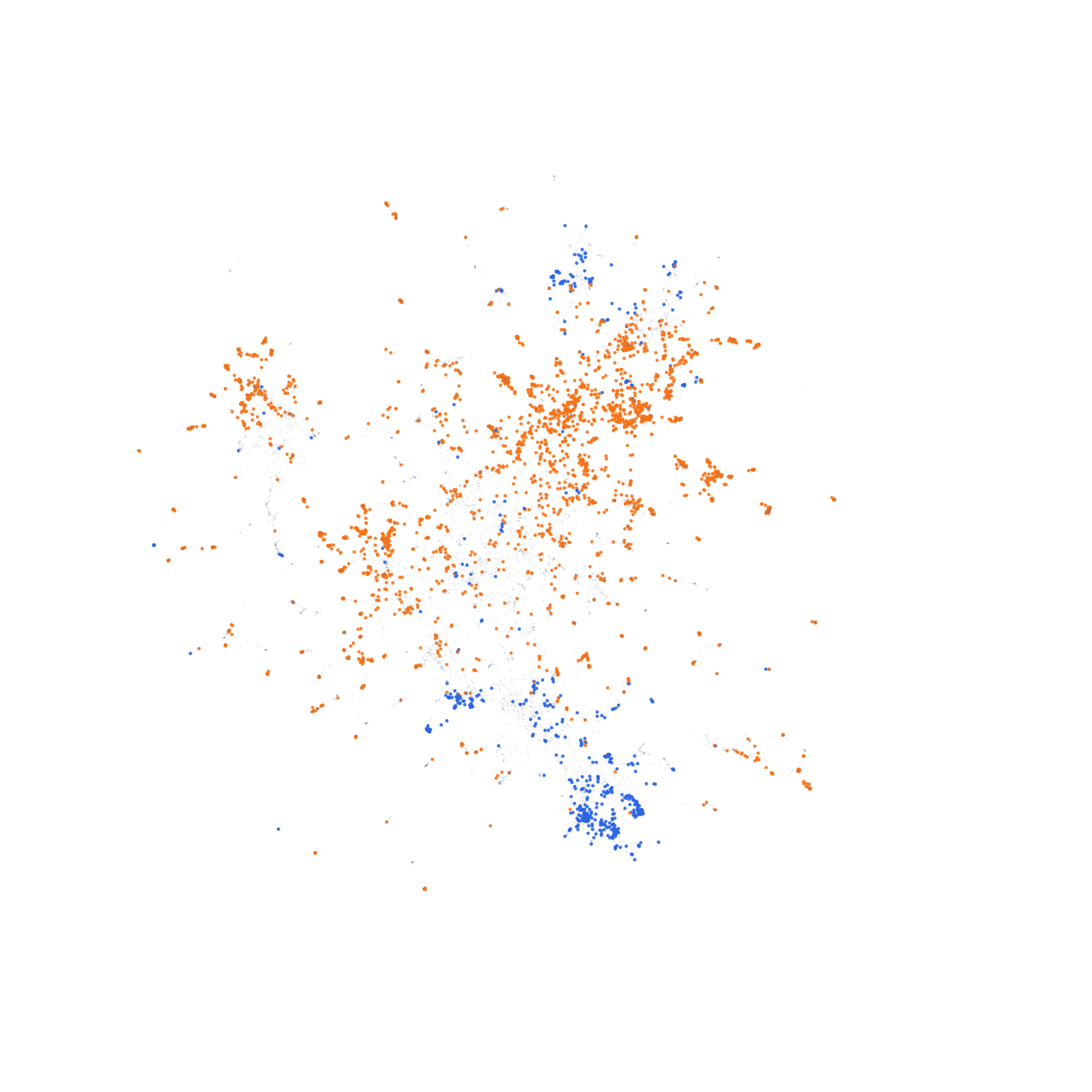}
\caption{Opportunity vs. Risk}
\end{subfigure}
\hfill
\begin{subfigure}[t]{0.32\linewidth}
\centering
\includegraphics[width=\linewidth]{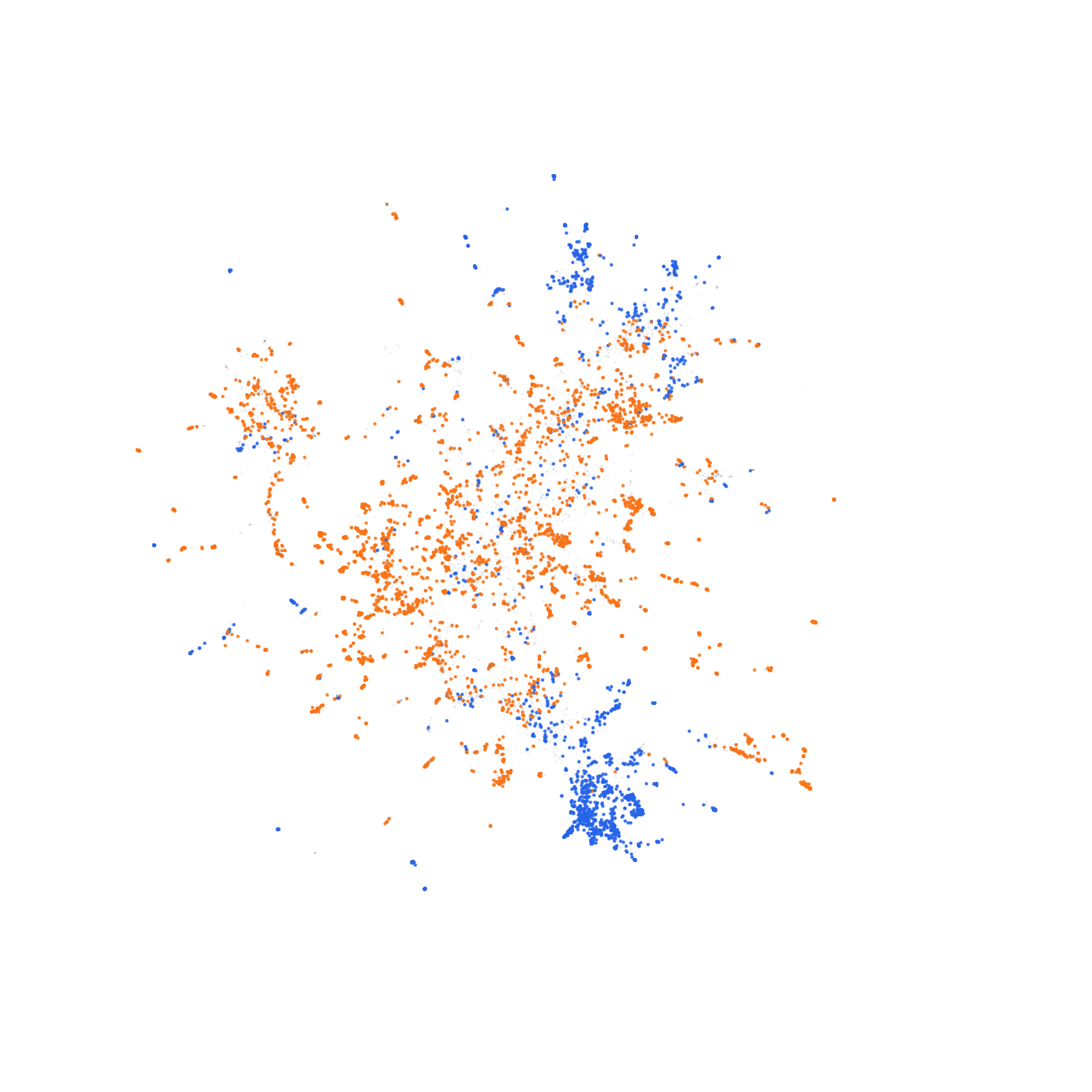}
\caption{Regulatory Pressure}
\end{subfigure}
\hfill
\begin{subfigure}[t]{0.32\linewidth}
\centering
\includegraphics[width=\linewidth]{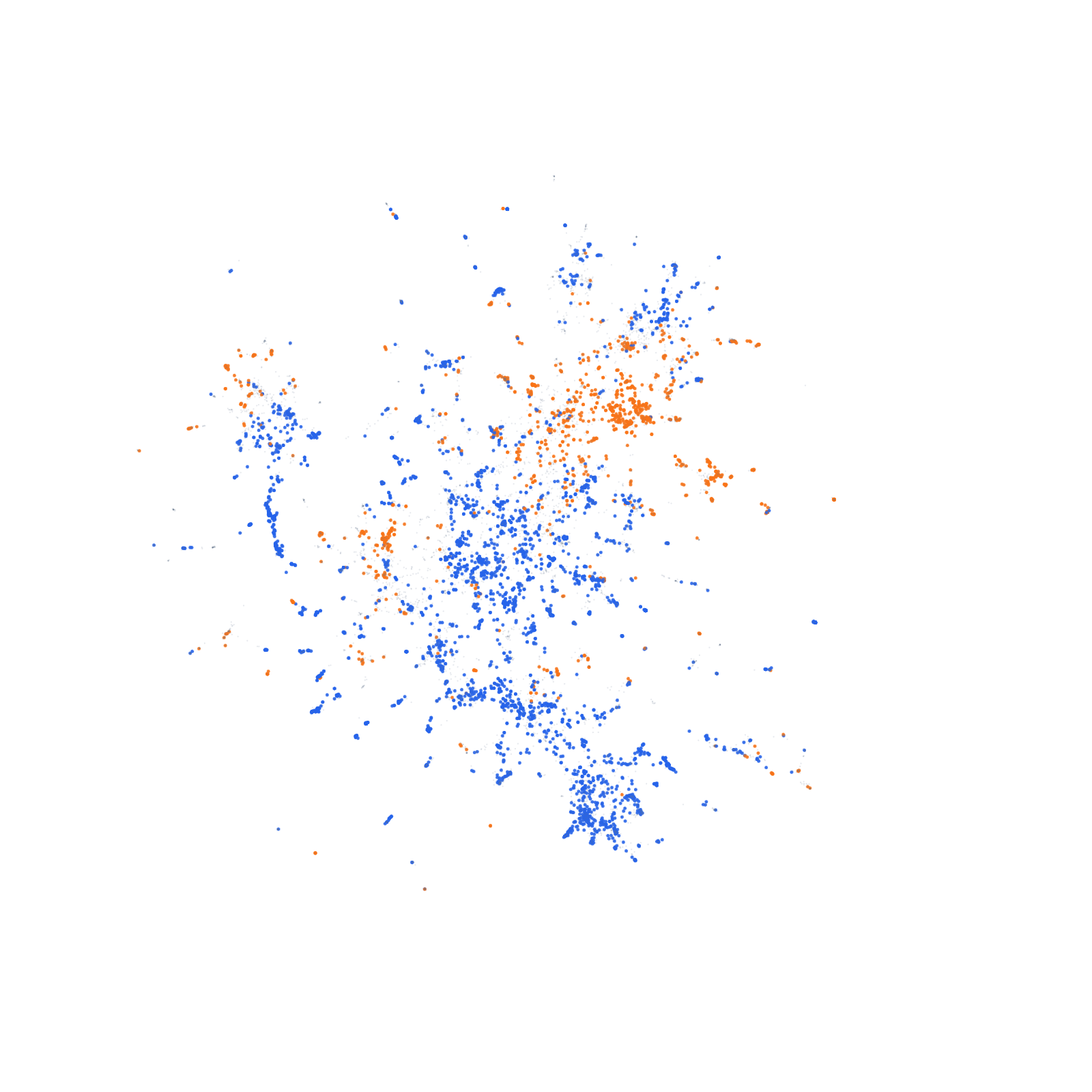}
\caption{Economic Momentum}
\end{subfigure}
\\[0.75em]
\begin{subfigure}[t]{0.32\linewidth}
\centering
\includegraphics[width=\linewidth]{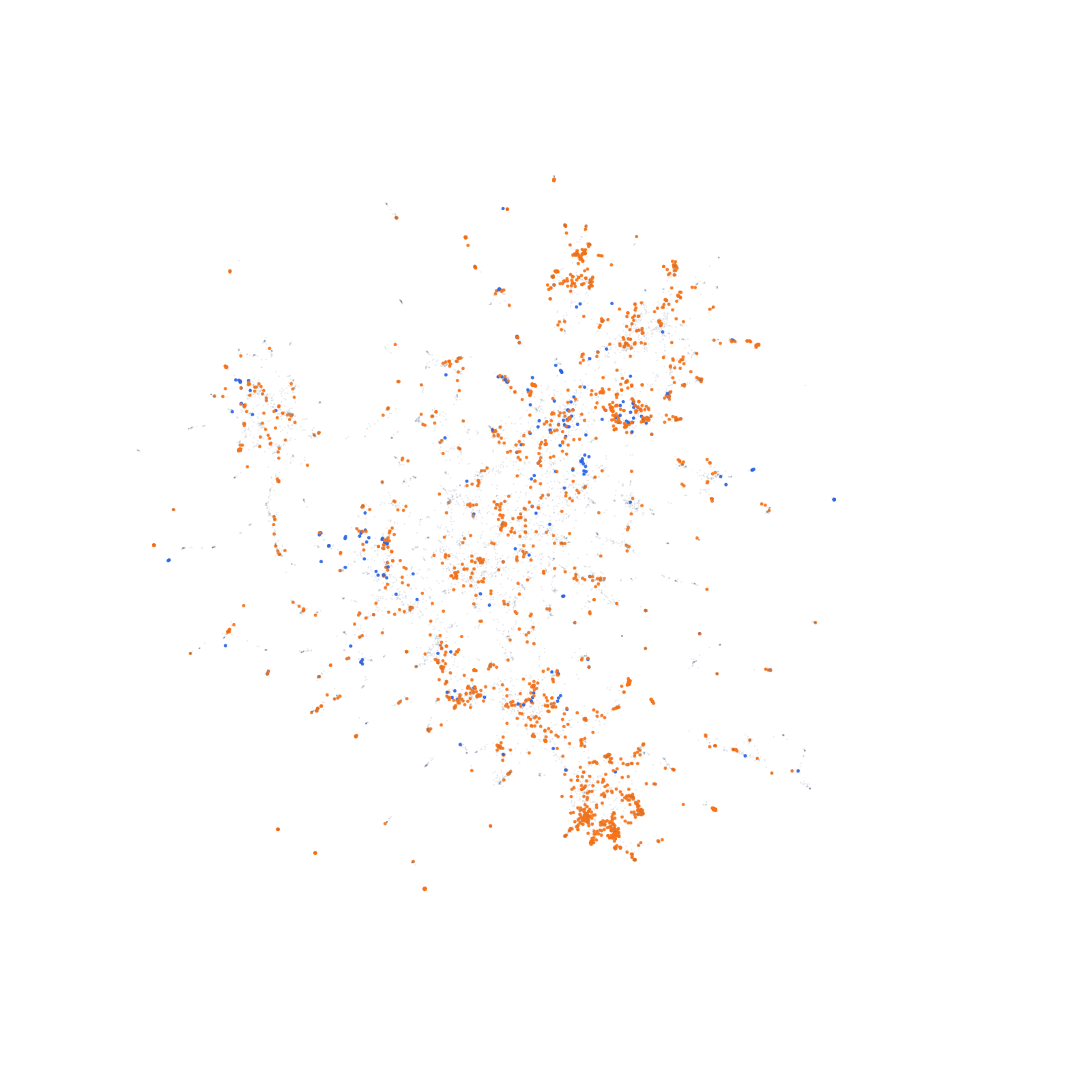}
\caption{Ethics vs. Utility}
\end{subfigure}
\hfill
\begin{subfigure}[t]{0.32\linewidth}
\centering
\includegraphics[width=\linewidth]{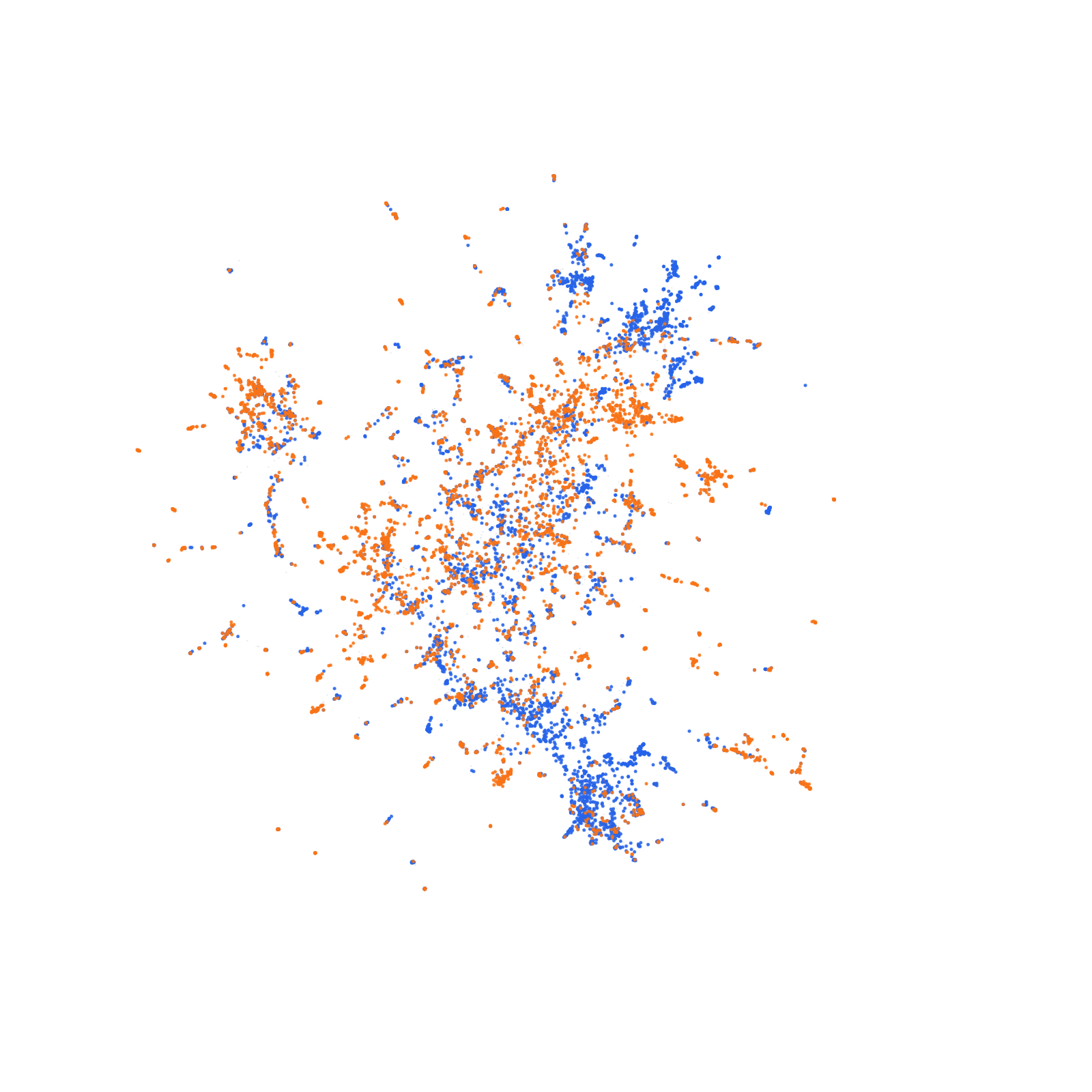}
\caption{Geopolitical Scope}
\end{subfigure}
\hfill
\begin{subfigure}[t]{0.32\linewidth}
\centering
\includegraphics[width=\linewidth]{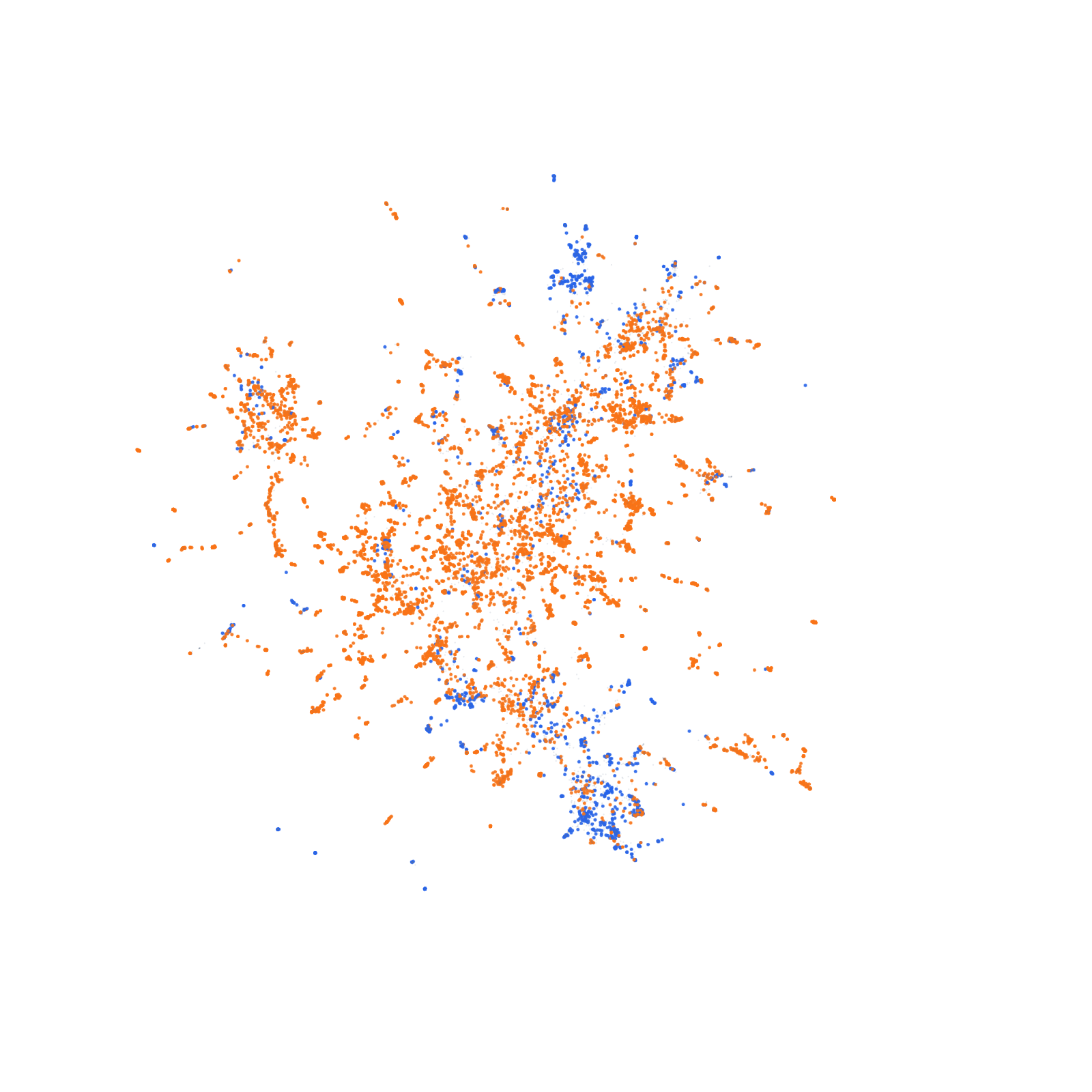}
\caption{Urgency}
\end{subfigure}
\caption{Representative article-level semantic identity dimensions projected onto the manifold. Across dimensions, opposite poles occupy different parts of the space rather than mixing uniformly across the corpus.}
\label{fig:short_semantic_identity_dimensions}
\end{figure}

Projecting these semantic identity dimensions back onto the noise-reduced map reveals a stronger structural fact (Fig.~\ref{fig:short_semantic_identity_dimensions}). Once the article-level logprob scores are painted onto the manifold, the opposite poles of each semantic dimension occupy distinct, localized regions. The logprob-derived poles are therefore not acting as superficial scalar tags placed on top of the map. They correspond strictly to recurrent spatial tendencies in the corpus, making it possible to read the spatial regions both geometrically and semantically.

\section{Limitations}

The paper documents a practical pipeline rather than a fully benchmarked methodological study. As a result, some of the design decisions are validated operationally inside the project but not through exhaustive comparison against alternative modeling choices.

The semantic indicators depend on prompt anchoring and model behavior. Although the logprob-based approach yields a coherent positional dictionary, this paper does not include prompt-sensitivity sweeps or human annotation of extreme scores.

Likewise, the structural map depends on practical choices such as $K=15$ for K-Means and the anomaly thresholds used in the pruning stages. These choices were validated operationally inside the project, but not through an exhaustive parameter study. In particular, the use of K-Means regions as part of the anomaly-detection procedure was effective in this pipeline, but it should still be treated as a practical design choice rather than a fully established methodological result, and it merits more systematic investigation.

Finally, this paper does not claim a formal supervised evaluation. Its goal is to document a compact and interpretable pipeline for semantic pattern recognition rather than to benchmark a classification system. The implementation is also tied to a high-throughput local inference stack built around \texttt{vLLM}, PostgreSQL, and GPU-backed Qwen models, so direct reproducibility depends on access to comparable infrastructure.
\section{Future Work}

Several natural extensions follow from the present pipeline. First, the same semantic identity profile can be indexed over time, allowing region-level and corpus-level trajectories to be studied through timestamps rather than only through static topography. This would make it possible to observe temporal drift, event shocks, and changes in barycentric position across the semantic dimensions.

Second, identity profiles derived from the positional dictionary may be useful as support variables in downstream learning settings. Aggregated over clusters or temporal windows, they could be tested as explanatory inputs in forecasting exercises or in models that relate semantic regions to external outcomes of interest.

Third, the observations removed during noise reduction should not necessarily be treated as disposable residuals. They may contain intrinsic analytical value, including edge cases, emergent weak signals, or structurally rare documents that deserve dedicated analytical processes rather than simple exclusion from the main map. %
\section{Conclusion}

We presented a practical pipeline for extracting quantitative semantic signals from text corpora. By combining full-document embeddings, article-level logprob-based scoring, manifold projection, structural segmentation, and anomaly detection for noise reduction, the workflow transforms unstructured text into operational, continuous data points.

The main value of this approach is operational. Each document receives a continuous semantic identity, and the corpus itself can be characterized through aggregated distributions over the same semantic dictionary. In practice, this turns text-as-signal processing into a concrete AI engineering capability for corpus inspection, automated monitoring, and downstream analysis, while reducing dependence on manually annotated labels for exploratory and support tasks. Just as importantly, the identity layer is not tied to a single universal schema: the semantic dictionary can be customized to the needs of a given analytical stream, making it possible to redefine the operative dimensions according to the monitoring, decision-support, or domain requirements of each use case. 

\printbibliography

\end{document}